\documentclass[10pt,twocolumn,letterpaper]{article}

\usepackage{wacv}
\usepackage{times}
\usepackage{epsfig}
\usepackage{graphicx}
\usepackage{amsmath}
\usepackage{amssymb}

\usepackage[pagebackref=true,breaklinks=true,letterpaper=true,colorlinks,bookmarks=false]{hyperref}

\wacvfinalcopy % *** Uncomment this line for the final submission

\def\wacvPaperID{****} % *** Enter the wacv Paper ID here

\ifwacvfinal\fi
\begin{document}

%%%%%%%%% TITLE
\title{Towards Learning a Self-inverse Network for Bidirectional Image-to-image Translation}

% Authors at the same institution
%\author{First Author \hspace{2cm} Second Author \\
% Authors at different institutions
\author{Zengming Shen \\
University of Illinois at Urbana-Champaign\\
{\tt\small zshen5@illinois.edu}
\and
Yifan Chen \\
University of Illinois at Urbana-Champaign\\
{\tt\small yifanc3@illinois.edu}
\and
S.Kevin Zhou \\
Medical Imaging, Robotics, Analystic Computing Laboratory \& Engineering (MIRACLE)\\
Institute of Computing Technology, Chinese Academy of Sciences\\
{\tt\small zhoushaohua@ict.ac.cn}
\and
Bogdan Georgescu \\
Siemens Healthineers\\
{\tt\small bogdan.georgescu@siemens-healthineers.com}
\and
Xuqi Liu \\
Rutgers University\\
{\tt\small xl325@scarletmail.rutgers.edu}
\and
Thomas S. Huang \\
University of Illinois at Urbana-Champaign\\
{\tt\small t-huang1@illinois.edu}
}

\maketitle
\ifwacvfinal\thispagestyle{empty}\fi

%%%%%%%%% ABSTRACT
\begin{abstract}
   The one-to-one mapping is necessary for many bidirectional image-to-image translation applications, such as MRI image synthesis as MRI images are unique to the patient. State-of-the-art approaches for image synthesis from domain X to domain Y learn a convolutional neural network that meticulously maps between the domains. A different network is typically implemented to map along the opposite direction, from Y to X. In this paper, we explore the possibility of only wielding one network for bi-directional image synthesis. In other words, such an autonomous learning network implements a self-inverse function. A self-inverse network shares several distinct advantages: only one network instead of two, better generalization and more restricted parameter space. Most importantly, a self-inverse function guarantees a one-to-one mapping, a property that cannot be guaranteed by earlier approaches that are not self-inverse. The experiments on three datasets show that, compared with the baseline approaches that use two separate models for the image synthesis along two directions, our self-inverse network achieves better synthesis results in terms of standard metrics. Finally, our sensitivity analysis confirms the feasibility of learning a self-inverse function for the bidirectional image translation.
\end{abstract}

%%%%%%%%% BODY TEXT
\section{Introduction}

\begin{figure}[t]
\begin{center}
   \includegraphics[width=\linewidth]{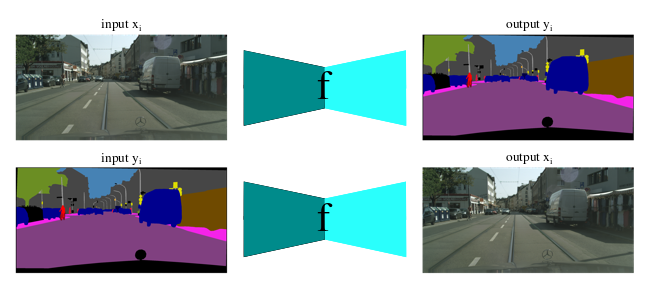}
\end{center}
   \caption{Our self-inverse network learns a bijective mapping $f: x_{i} \leftrightarrow y_{i} $.Here we illustrate the concept using the CityScapes dataset \cite{cordts2016cityscapes} for bidirectional photo-to-label translation.}
\label{fig:long}
\label{fig:onecol}
\end{figure}

Recently there is a growing need for bidirectional image-to-image translation, include image style transfer, translation between image and semantic labels, gray-scale to color, edge-map to photograph, super resolution and many other types of image manipulations. Here we highlight one application related to Magnetic Resonance Imaging (MRI). MRI is one of the widely used medical image modalities due to its non-invasiveness and its ability of clearly capturing soft tissue structures using multiple acquisition sequences. However, its disadvantage lies in its long acquisition time and expensive cost. Therefore, there is a lack of large scale MRI image database needed for learning-based image analysis. MRI image synthesis or image-to-image translation~\cite{xiang2018ultra,huang2017simultaneous} is able to fill such a gap by generating more images for training purpose. Also, a generated MRI image can be helpful to cross-sequence image registration, in which an image is first synthesized for the target sequence and then used for registration~\cite{chen2015using}. 

In language translation, if we treat the translation from one language A to another language B as a forward process $f$, then the translation from the language B to A is its inverse problem $f^{-1}$. Similarly, in computer vision, there is a concept of image-to-image translation~\cite{pix2pix2016,CycleGAN2017,zhu2017toward,choi2018stargan} that converts an image to another one. In medical imaging, there are image reconstruction problems. Traditionally, each of these problems uses two different functions, one for the forward task $f$ and the other one for its inverse $f^{-1}$. In this paper, our goal is to demonstrate that, for MRI image synthesis and other tasks, we are able to {\it learn the above two tasks simultaneously using only one function} (see
Figure 1), that is, $f=f^{-1}$.

The community has explored the power of CNN in various tasks in computer vision, as well as within several other fields. But so far, to the best of our knowledge, no one has explored the learning capability of a self-inverse function using CNN, and its potential use in applications. Our aim in this paper is to bridge this gap. We refer to the mapping from a domain $X$ to a domain $Y$ as task $A$ and the mapping from the domain $Y$ to $X$ as task $B$. Additionally, the proposed CNN that learns a self-inverse function is referred to as the self-inverse or one-to-one network. The one-to-one mapping property is necessary for application like MRI image synthesis as MRI images are unique to the patient.

\begin{figure}[t]
\begin{center}
   \includegraphics[width=0.9\linewidth]{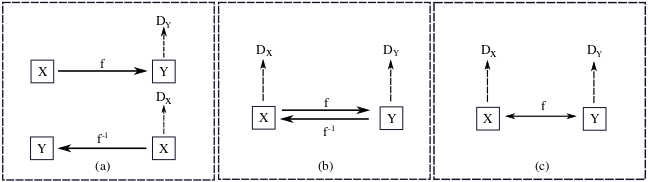}
\end{center}
   \caption{A comparison of our self-inverse network and other CNNs for image-to-image translation. The $f$ and $f^{-1}$ are the two generator networks for the tasks $A$ and $B$, respectively. The $D_{Y}$ and the $D_{X}$ are the associated adversarial discriminators. (a) Pix2pix \cite{pix2pix2016}: Two separate generators networks $f$ and $f^{-1}$ for the tasks $A$ and $B$, respectively. (b) Cycle GAN \cite{CycleGAN2017}: Two jointly trained but different generator networks $f$ and $f^{-1}$ for the tasks $A$ and $B$, respectively. (c) Self-inverse network: Only one generator network for both tasks.}
\label{fig:long}
\label{fig:onecol}
\end{figure}

\begin{figure}[t]
\begin{center}
   \includegraphics[width=0.9\linewidth]{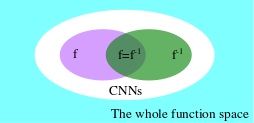}
\end{center}
   \caption{Function space. Blue area: the whole function space;White area:the function space of a CNN; Purple area: the function space of $f$; Green area: the function space of $f^{-1}$; and Overlap area: the function space of $f = f^{-1}$.}
\label{fig:long}
\label{fig:onecol}
\end{figure}

\section{Benefits of learning a self-inverse network}

There are several advantages in learning a self-inverse network equipped with the one-to-one mapping property.

(1) From the perspective of the application, only one self-inverse function can model both tasks $A$ and $B$ and it is a novel way for multi-task learning. As shown in Figure 1, the self-inverse network generates an output given an input, and vice versa, with only one CNN and without knowing the mapping direction. It is capable of doing both tasks within the same network, simultaneously. In comparison to separately assigning two CNNs for tasks $A$ and $B$, the self-inverse network halves the necessary parameters, assuming that the self-inverse network and the two CNNs share the same network architecture as shown in Figure 2.

(2) It automatically doubles the sample size, a great feature for any data-driven models, thus becoming less likely to over-fit the model. The self-inverse function $f$ has the co-domain $ Z = X \cup Y$. If the sample size of either domain $X$ or $Y$ is $N$, then the sample size for domain $Z$ is $2N$. As a result, the sample size for both tasks $A$ and $B$ are doubled, becoming a novel method for data augmentation to mitigate the over-fitting problem. 

(3) It implicitly shrinks the target function space. As shown in Figure 3, the blue area is the whole function space, which is unlimited. Given a CNN with its architecture fixed, its function space (Figure 3, white area) is enormous, with millions of parameters. When the CNN is trained for the task $A$, the target function space $f$ is the purple area. When the CNN is trained for the task $B$, the target function space $f^{-1}$ is the green area. When it is trained to learn a self-inverse function for both tasks $A$ and $B$, the target function space is the overlapping area, which is a subset of the function space of $f$ and $f^{-1}$. For a fixed neural network architecture, its function space is large enough to have the overlapping area in Figure 3. For a fixed data set, the trained model is a function within the blue area or the purple area for each direction, since the overlap area is always the subset of the blue or purple areas. If the network is trained as a self-inverse network, the trained model is a function within the overlapping area, which is always smaller than that of the network trained separately in each direction. A smaller function space means a smaller bias between the true function and the trained model, so the self-inverse network likely generalizes better. Another interpretation of this shrinking behavior is to regard the inverse $f^{-1}$ as a regularization condition when learning the function $f$, and vice versa.
 
\begin{figure}[t]
\begin{center}
   \includegraphics[width=0.95\linewidth]{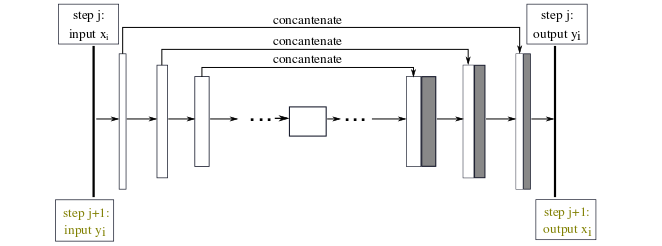}
\end{center}
   \caption{The illustrations of the self-inverse network using the U-Net architecture \cite{ronneberger2015u}. Each block represents the Convolution-BatchNorm-LeakyReLU layers  in the encoder part and the Convolution-BatchNorm-ReLU layers in the decoder. Alternative training: In the training stage, for a batch of image pairs $(x_{i},y_{i})$, at the step j, the input and label are $x_{i}$ and $y_{i}$, respectively, at the step $j+1$, the input and label are $y_{i}$ and $x_{i}$, respectively.}
\label{fig:long}
\label{fig:onecol}
\end{figure}

\begin{table}
\begin{center}
\begin{tabular}{|l|c|c|c|c|r|}
\hline
Direction&Method & p. acc.$\uparrow$ & c. acc.$\uparrow$ & IOU$\uparrow$ \\
\hline\hline
photo$\rightarrow$label&pix2pix    &0.80&0.35& 0.29\\
\hline
photo$\rightarrow$label&one2one     &\textbf{0.83}&0.35& 0.29\\
\hline\hline
label$\rightarrow$photo&pix2pix    &0.73&0.25& 0.19\\
\hline
label$\rightarrow$photo&one2one     &\textbf{0.74}&0.25& \textbf{0.20}\\
\hline\hline
labels$\rightarrow$photo&GT      &0.80&0.26& 0.21\\
\hline
\end{tabular}
\end{center}
\caption{Quantitative performance of labels$\leftrightarrow$photo on cityscapes dataset.}
\end{table}

\begin{figure}[t]
\begin{center}
\centerline{\includegraphics[width=\linewidth]{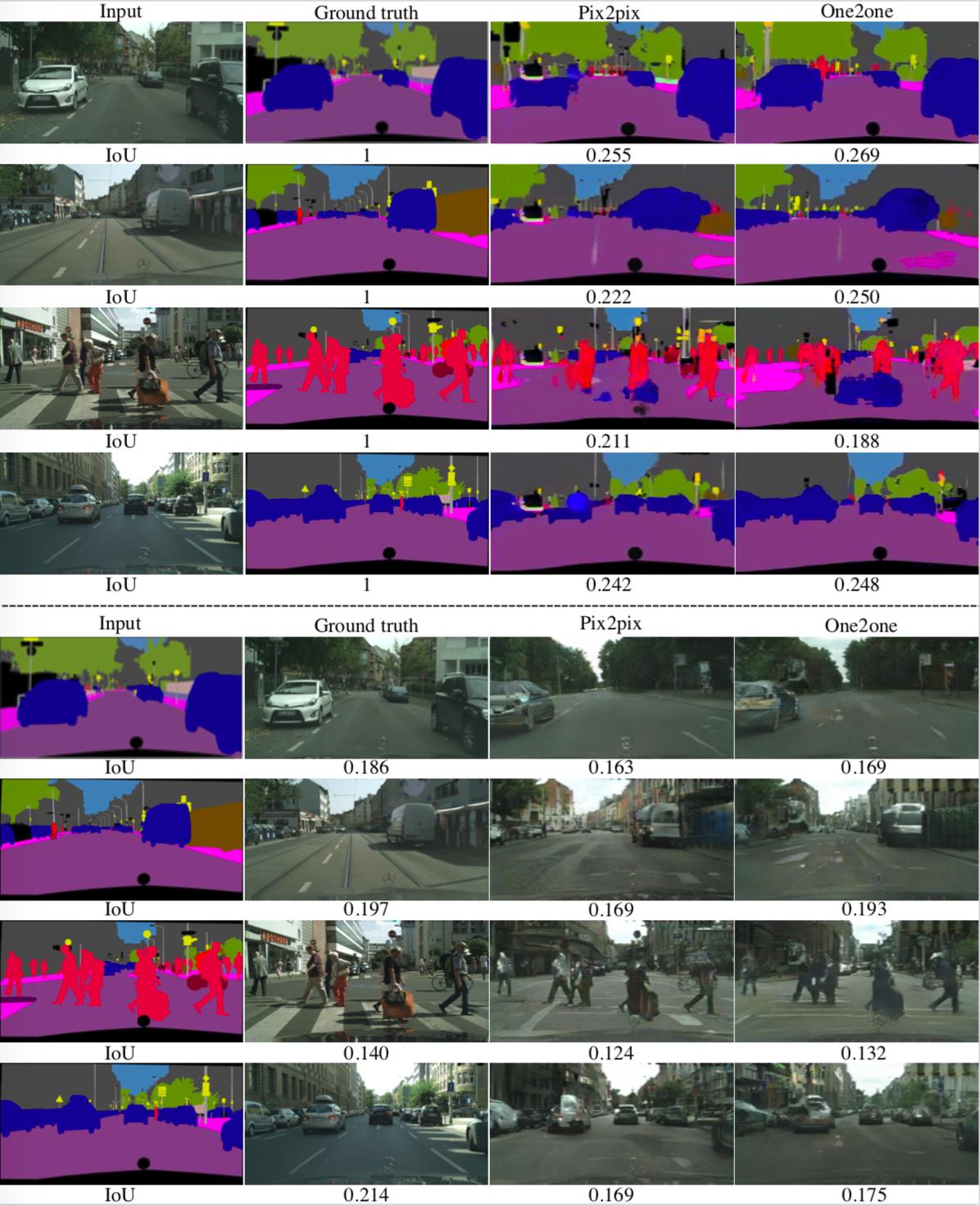}}
\end{center}
   \caption{Qualitative result on labels$\leftrightarrow$photo bidirectional image-to-image translation on cityscapes dataset. Upper: photo $\rightarrow$ label. Bottom: label $\rightarrow$ photo.}
\label{fig:long}
\label{fig:onecol}
\end{figure}

%-------------------------------------------------------------------------
\section{Related Works}

\textbf{Inverse problem with neural networks}
The loss of information is a big problem that affects the performance of CNNs in various tasks. Several works such as \cite{dosovitskiy2016inverting,mahendran2016visualizing} show that essential information concerning the input image is lost as the network traverses to deeper layers in well-known ImageNet-based  CNN classifiers. To recover and understand the loss of information, the above works use learned or hand-crafted methods prior to inverting the representation. An example of `compensating' the lost information for performance improvement involves the segmentation task approach \cite{dalca2018anatomical}, which proposes the use of prior anatomical information from the latent space within a pre-trained decoder.

Building an invertible architecture is difficult due to the local inversion being ill-conditioned,  %Several works demonstrate that building an invertible CNN is a difficult challenge, 
hence not much progress has been made in solving it. Multiple works only allow invertible representation learning under certain conditions. 
Parseval network \cite{cisse2017parseval} increases the robustness of learned representation with respect to adversarial attacks. In this work, the linear operator is bijective under the condition that the spectrum of convolutional operator is constrained to norm 1 during learning. \cite{bruna2013signal} introduces a signal recovery method conditioned on pooling representation to design invertible neural network layers. \cite{jacobsen2018revnet} makes the CNN architecture invertible by providing an explicit inverse. In this work, the reconstruction of the linear interpolations between natural image representation is achieved. This gives empirical evidence to the notion that learning invertible representation that do not discard any information concerning their input on large-scale supervised problems is possible. But it can not provide bi-directional mapping and is not self-invertible. Ardizzone et.al\cite{ardizzone2018analyzing}  prove theoretically and verify experimentally for artificial data and real data in inverse problme using invertible neural networks. More specifically, Kingma~\cite{kingma2018glow} uses the invertible 1x1 convolution for the generative flow. Different from the previous works, our self-inverse network realize the inevitability between two domains by switching the inputs and outputs and then learning a self-inverse function.

\begin{table}[t]
\begin{center}
\begin{tabular}{|l|c|c|c|c|r|}
\hline
Direction& Method & L1$\downarrow$ & PSNR$\uparrow$ & SSIM $\uparrow$\\
\hline\hline
aerial$\rightarrow$map&pix2pix   &0.0696&19.36& 0.505  \\
\hline
aerial$\rightarrow$map&one2one     & \textbf{0.0635}& \textbf{19.93} & \textbf{0.558}\\
\hline
map$\rightarrow$aerial&pix2pix    & 0.270 & 9.091& 0.144  \\
\hline
map$\rightarrow$aerial&one2one     & 0.270& \textbf{9.101} &\textbf{0.148}\\
\hline
\end{tabular}
\end{center}
\caption{Quantitative performance of map$\leftrightarrow$aerial on google maps.}
\end{table}

\begin{figure}[t]
\begin{center}
   \includegraphics[width=0.8\linewidth]{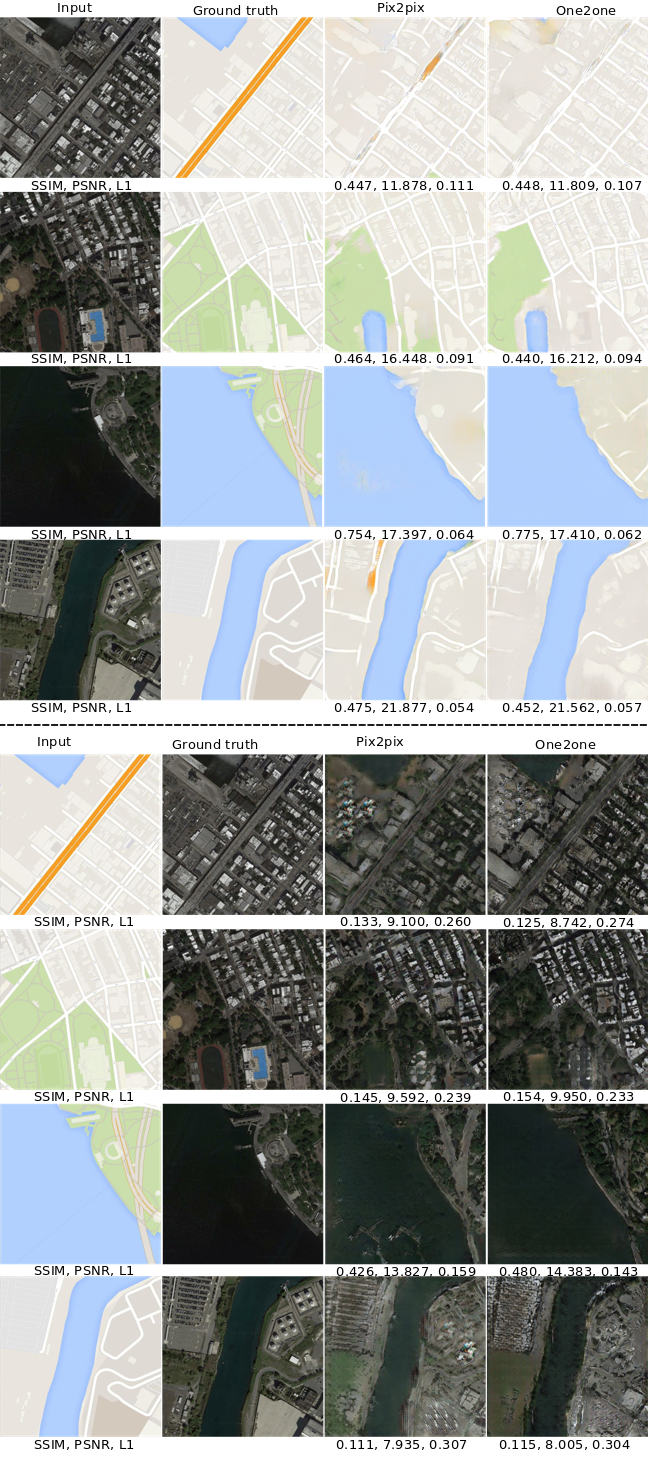}
\end{center}
   \caption{Qualitative result of bidirectional image-to-image translation on aerial$\leftrightarrow$map on google maps. Upper: earial$\rightarrow$map. Buttom: map $\rightarrow$ aerial.}
\label{fig:long}
\label{fig:onecol}
\end{figure}

\textbf{Image-to-image translation} 
The concept of image-to-image translation is broad, including image style transfer, translation between image and semantic labels, gray-scale to color, edge-map to photograph, super resolution \cite{liu2016robust} and many other types of image manipulations. It dates back to image analogies by \cite{hertzmann2001image}, which employs a non-parametric texture model \cite{efros1999texture} from a single input-output training image pair. More recent approaches use a data set of input-output examples to learn a parametric translation function using CNN \cite{long2015fully}. Our approach builds on the “pix2pix” framework of \cite{pix2pix2016}, which uses a conditional generative adversarial network \cite{goodfellow2014generative} to learn a mapping from input to output images. CycleGAN \cite{CycleGAN2017} contributes to the unpaired image-to-image translation with a cycle consistency loss. In this framework, CycleGAN addresses exactly the same issue of learning a bijective mapping, albeit without the self-inverse property. CycleGAN can be seen as BiGAN \cite{donahue2016adversarial} where the latent variable is like an image in the co-domain and the loss is augmented with an L1 loss. Similar ideas have been applied to various tasks such as generating photographs from sketches \cite{sangkloy2016scribbler} or from attribute and semantic layouts\cite{karacan2016learning}. Recently, ~\cite{wang2018pix2pixHD} uses multi-scale loss and Conditional GAN to realize high resolution image synthesis and semantic manipulation. One direction towards diverisifying image translation is to allow many to many mapping, like augmented CycleGAN\cite{almahairi2018augmented,liu2017unsupervised,lee2018diverse,huang2018multimodal,zhu2017toward,lee2019drit++}. The other direction towards accurate image translation is to restrict output image variance, like instance level image translation~\cite{shen2019towards}. Our method falls into the latter case and  learns both tasks $A$ and $B$ with one generator network in a bidirectional way instead of using two generator networks (see Figure 2).Unlike \cite{zhu2017toward}, we encourage the invertbility of our model as a self-inverse function to realize bijection.

\textbf{Neural style transfer}
Neural style transfer can be treated as a special category of image-to-image translation as well. \cite{gatys2016image} proposes to use image representation derived from CNN, optimized for object recognition, to make high level image information explicit. \cite{chen2017photographic} introduces a cascade refinement networks for photographic image synthesis. \cite{ulyanov2016texture} highlights the power
and flexibility of generative feed-forward models
trained with complex and expressive loss functions for style transfer.  \cite{johnson2016perceptual} contributes the perceptual losses, which works very well.

\section{Method}

Our goal is to learn a self-inverse mapping function or bidirectional mapping function $f$ for pairs $(x_{i},y_{i})$. This means $f: x_{i} \leftrightarrow y_{i}$. It also can be illustrated in this way: the function $f: x_{i} \rightarrow y_{i}$ and its inverse function $f^{-1}: y_{i} \rightarrow x_{i}$ satisfies $ f = f^{-1} $, where samples  $\left \{x_{i}  \right \}_{i=1}^N \in X $, $\left \{y_{i}  \right \}_{i=1}^N \in Y $, and the symbol `$\leftrightarrow$' means bijection: the symbol `$\rightarrow$' means one directional mapping and the symbol `$=$' means the two functions on both sides are exactly the same function.

\begin{table}[t]
\begin{center}
\begin{tabular}{|c|l|c|r|}
\hline
Direction&Method &  d(Class IOU)$\uparrow$ \\
\hline\hline
labels$\rightarrow$photo&pix2pix    & 0.0168\\
\hline
labels$\rightarrow$photo&one2one     & \textbf{0.0178}\\
\hline\hline
photo$\rightarrow$labels&pix2pix    & \textbf{0.0199}\\
\hline
photo$\rightarrow$labels&one2one     & 0.0190\\
\hline
\end{tabular}
\end{center}
\caption{Model sensitivity performance of labels$\leftrightarrow$photo on cityscapes.}
\end{table}

Mathematically, it boils down to solving the following minimization problem:
\begin{equation}
  \min_W \sum_{i=1}^N l_A( f_W(x_i), y_i) + l_{B}(x_i,  f_W(y_i)) + \lambda~r(W),
\end{equation}
where $W$ denotes the neural network parameters, $l_A$ and $l_B$ the loss function for tasks $A$ and $B$, respectively, and $r(W)$ is the regularizer. In this paper, we use $L_1$ norm as the loss and GAN discriminator as the regularizer. The model pipeline is illustrated in Figure 2(c). It consists of two networks. The generator network $f$ and the discriminator network $D_{x}$ or $D_{y}$. Here $D_{x}$ and $D_{y}$ are the same network, while the $D_{x}$ and $D_{y}$ are two different networks for the baseline pix2pix model (see Figure 2(a)). The generator $f$ is trained to translate the image as real as possible to fool the discriminator network $D_{x}$ or $D_{y}$, which is trained as well as possible to detect the `fake' examples generated by $f$.

%-------------------------------------------------------------------------

\begin{figure}[t]
\begin{center}
   \includegraphics[width=0.8\linewidth]{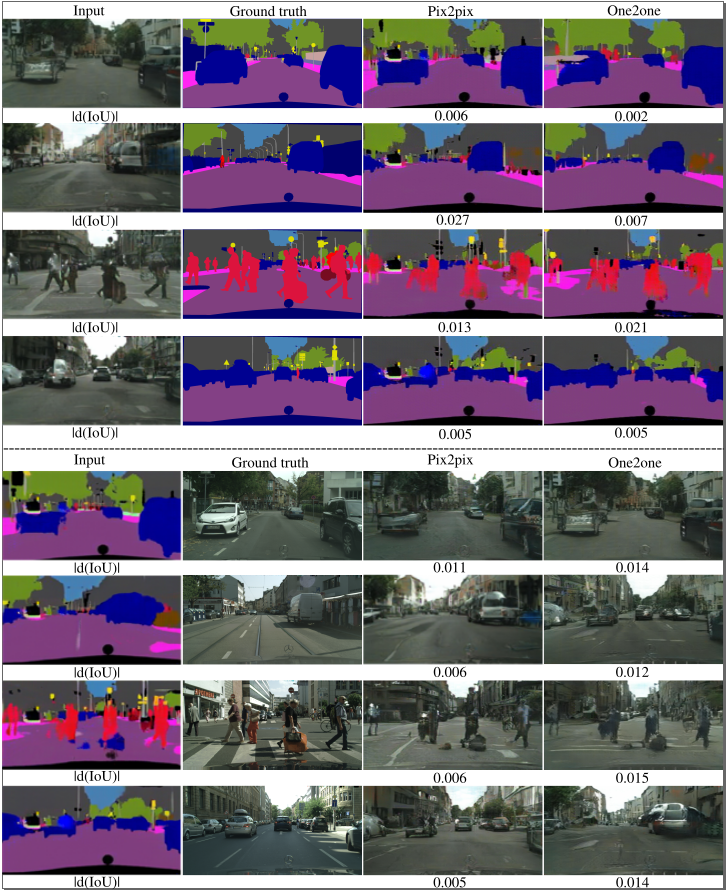}
\end{center}
   \caption{Model sensitivity performance of labels$\leftrightarrow$photo on cityscapes. Upper: photo $\rightarrow$ labels. The input is generated by inputting the groudtruth  to pix2pixB. Bottom: labels $\rightarrow$ photo. The input is generated by inputting the groudtruth to pix2pixA.}
\label{fig:long}
\label{fig:onecol}
\end{figure}

\textbf{Detailed network architecture.} We adopt the architecture from~\cite{pix2pix2016} for our self-inverse network implementation. Let $C_k$ denote a Convolution-BatchNorm-LeakyReLU layer with $k$ filters in the encoder and Convolution-BatchNorm-ReLU layer with $k$ filters in the decoder.  All convolutions are $4 \times 4 $ spatial filters applied with a stride 2. Convolutions in the encoder are down-sampled by a factor of 2. Convolutions in the decoder are up-sampled by a factor of 2.

The encoder-decoder architecture consists of an encoder,
$C_{64}-C_{128}-C_{256}-C_{512}-C_{512}-C_{512}-C_{512}-C_{512}$,
and an decoder, $C_{512}-C_{512}-C_{512}-C_{512}-C_{512}-C_{256}-C_{128}-C_{64}$.
After the last layer in the decoder, a convolution is applied to map according to the number of output channels, which is 1, followed by a Tanh function. Following the convention, The $C_{64}$ is not applied with batch-normalization. All LeakyReLUs in the encoder are with a slope of 0.2. For the U-Net skip connection, the skip connection is to concatenate feature maps from layer $i$ to layer $n-i$. where $i$ is the layer index, $n$ is the total number of layers. Compared to the decoder above without skip connection, the number of feature maps doubles due to the use of an U-Net decoder,
$C_{512}-C_{1024}-C_{1024}-C_{1024}-C_{1024}-C_{512}-C_{256}-C_{128}$.It is
$C_{64}-C_{128}-C_{256}-C_{512}$. Following the $C_{512}$ layer is a convolution layer to map the feature map channel number to 1. Then a sigmoid function is followed to generate the output. Similar to the generator, the first convolution layer $C_{64}$ is without batch normalization. All LeakyReLU are with a slope of 0.2.

\begin{table}
\begin{center}
\begin{tabular}{|l|c|c|c|c|r|}
\hline
Direction&Method & dL1$\downarrow$ & dPSNR$\uparrow$ & dSSIM$\uparrow$ \\
\hline
aerial$\rightarrow$map&pix2pix    & .0007& 0.87 & 0.029 \\
\hline
aerial$\rightarrow$map&one2one     &\textbf{.0008}&\textbf{0.89}& 0.029\\
\hline\hline
map$\rightarrow$aerial&pix2pix    & 0.0140 & 0.447& 0.023  \\
\hline
map$\rightarrow$aerial&one2one    & \textbf{0.0144}& \textbf{0.458} &\textbf{0.024}\\
\hline
\end{tabular}
\end{center}
\caption{Model sensitivity performance of aerial$\leftrightarrow$map on Maps dataset.}
\end{table}

\textbf{Loss function.} 
The objective of a conditional GAN~\cite{mirza2014conditional} can be expressed as
        \begin{equation}
         \mathcal{L}_{cGAN}(G,D) = \mathop{\mathbb{E}_{x,y}}[\log D(x,y)]+\mathop{\mathbb{E}_{x,z}}[\log(1-D(x,G(x,z)))]
        \vspace{-0.5em} 
    \end{equation}
We use L1 distance rather than L2 as
L1 encourages less blurring:
    \begin{equation}
         \mathcal{L}_{L1}(G) = \mathop{\mathbb{E}}_{x,y,z}[||y-G(x,z)||_1]
        \vspace{-0.5em} 
    \end{equation}
Our final objective
\begin{equation}
        (G^*,D^*) = arg \mathop{min}_{\mathbf{G}} \mathop{max}_{\mathbf{D}}
        \mathcal{L}_{cGAN}(G,D) + \lambda\mathcal{L}_{L1}(G)
        \vspace{-0.5em}
    \end{equation}
With $z$, the net could learn a mapping from $x$ to $y$ in term of any distribution instead of just a delta function.

\begin{figure}[t]
\begin{center}
   \includegraphics[width=0.8\linewidth]{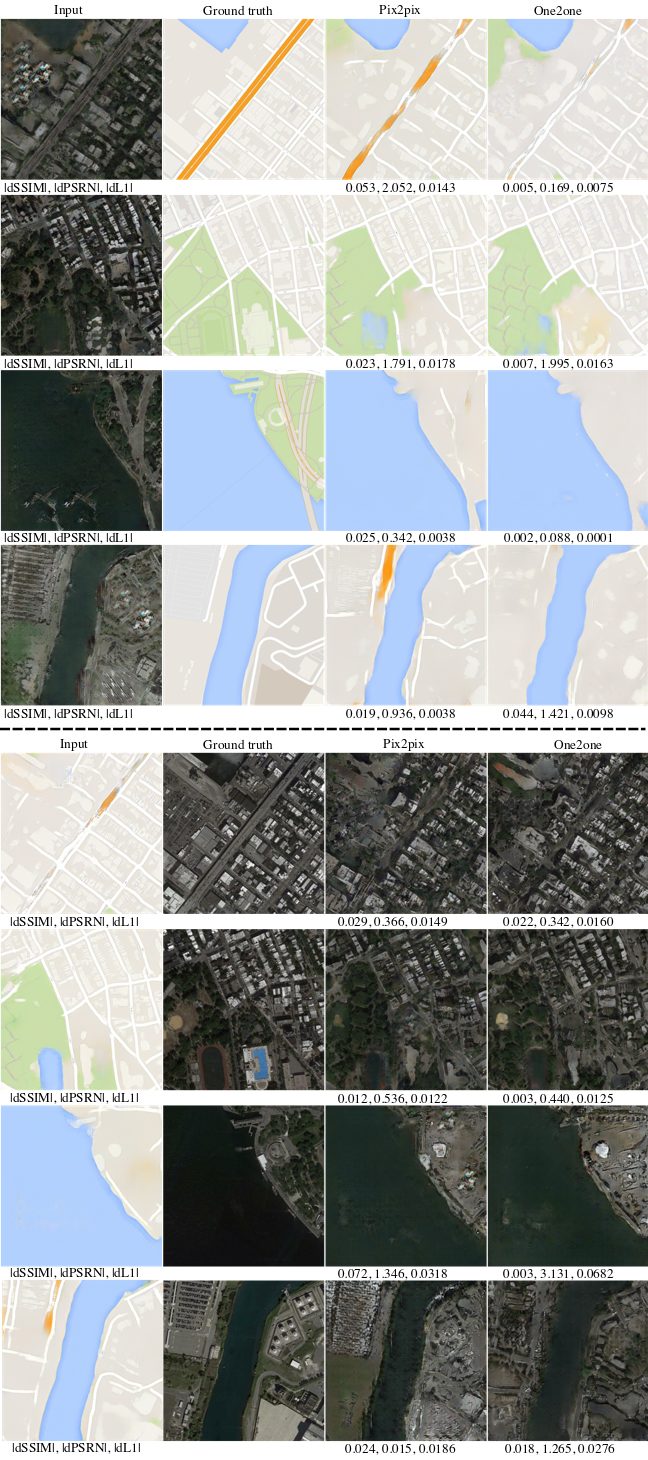}
\end{center}
   \caption{Example of caption.  It is set in Roman so that mathematics
   (always set in Roman: $B \sin A = A \sin B$) may be included without an
   ugly clash.}
\label{fig:long}
\label{fig:onecol}
\end{figure}

\textbf{Bi-directional Training} To train a CNN as a self-inverse network, we randomly sample a certain-sized batch of pairs $(x_{i},y_{i})$ and $(y_{i},x_{i})$ alternatively and iteratively. This is shown in (see
Figure.4). %In the bracket, the first item and the second item feed as the input and the ground truth, respectively. We refer to this as alternative training. 
The baseline is without alternative training, which means that training two separated generator networks for the tasks $A$ and $B$, respectively (see
Figure.4). For a fair comparison with the baseline, with the same data set, we use the same batch size and the same number of epochs. In other words, except for the alternative part, everything is the same as the baseline. We resize the $256\times 256$ input images to $286\times 286$, add a random jitter, and then randomly crop it back to size $256\times 256$. All networks are trained from scratch. The weights are initialized from a Gaussian distribution with mean 0 and standard deviation of 0.02. % All models a
\section{Experimental results}
Below, `pix2pix' refers to the result obtained by the model we retrained from scratch following exactly the same training details as that in the pix2pix paper ~\cite{pix2pix2016}. `one2one' refers to our results by training the same networks as a self-inverse function.
In all the tables, all of the results are averaged across the whole validation partition which follow the same dataset split in ~\cite{pix2pix2016} . %For Figures 5-8, the number below each image corresponds to the image above it individually. 

We conduct the experiments using three paired image data sets:
\begin{itemize}
    \item {Semantic label $\leftrightarrow$ photo}, trained on the Cityscapes dataset~\cite{cordts2016cityscapes};
    \item {Map $\leftrightarrow$ aerial photo}, trained on data scraped from
Google Maps ~\cite{pix2pix2016};
    \item {MRI image synthesis on BRATS}.
\end{itemize}
We use the following {evaluation metrics}
\begin{itemize}
\item  Cityscapes data set\cite{cordts2016cityscapes}. For fair comparison with the baseline, which is pix2pix ~\cite{pix2pix2016}, we follow the same evaluation metric as that in pix2pix paper. We use the released public evaluation code from the ~\href{https://github.com/phillipi/pix2pix/tree/master/scripts/eval_cityscapes}{pix2pix GitHub repository}. 
For the photo$\rightarrow$labels direction, we use IOU as the evaluation metric.
For the labels$\rightarrow$photo direction, we use the "FCN score" \cite{salimans2016improved,long2015fully,wang2016generative,zhang2016colorful,owens2016visually}.

%\item Architectural labels $\leftrightarrow$ photo, trained on the CMP Facades dataset~\cite{tylevcek2013spatial}.
\item Map data scraped from
Google Maps\cite{pix2pix2016} and Brats\cite{brats1}. To quantify the image quality distance between the generated image and the ground truth objectively and to have a metric to do the model sensitivity analysis, we use the SSIM\cite{wang2004image}, PSNR\cite{hore2010image}, and L1 distance as the evaluation metric for both directions.
\end{itemize}
% \subsection{Feature map and model parameter visualization}
% Due to the page limit, please refer to supplementary file.

\subsection{Semantic label $\leftrightarrow$ photo}

Our model is one2one and the baseline is pix2pix. Table 1 and Figure 5 show the model performance comparison between one2one model and pix2pix model on bidirectional label and photo image translation.
The evaluation metrics are pixel actuary(p.acc.), class accuracy(c.acc.) and class IOU(IOU). In the direction of photo $\rightarrow$ labels, our one2one model performances higher than pix2pix model by 3.75\% in pixel actuary. In the direction of labels $\rightarrow$ photo, the evaluation metric is ``FCN score". Our one2one model increase the class IOU by 5.3\% compared with the pix2pix model. Note that the FCN score for ground truth is 0.21. The FCN score of The one2one model is 0.20 which is very close to the score of the ground truth. 

\begin{figure*}[t]
\begin{center}
   \includegraphics[width=0.8\linewidth]{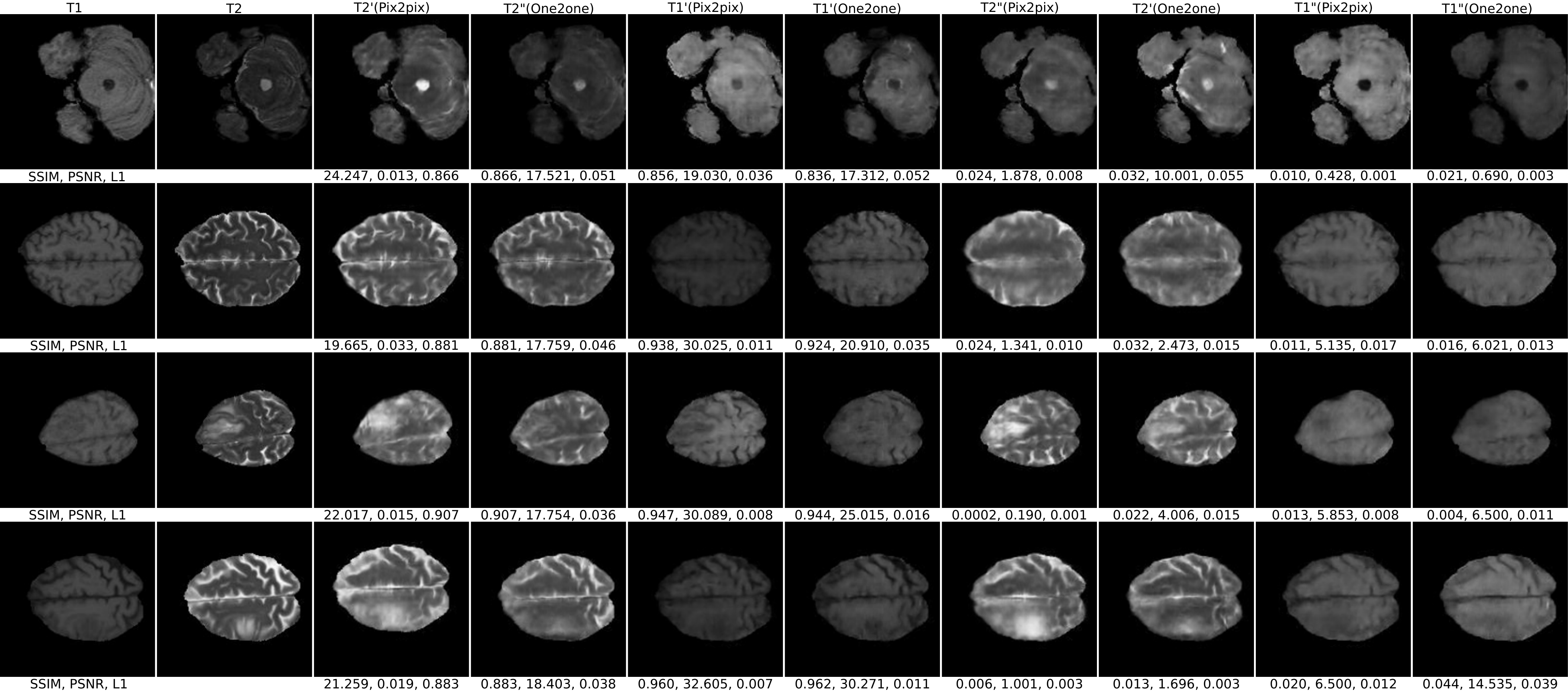}
\end{center}
   \caption{Examples of generated images. The column 1 depicts the original images for $T_1$. The column 2 depicts the original images for $T_2$. Generated $T_2$ images from $T_1$ with pix2pix and one2one models are in the column 3 and 4 respectively. Generated $T_1$ images from $T_2$ with with pix2pix and one2one models are in the column 5 and 6 respectively. Generated $T_2$ images from column 5 with pix2pix and one2one models are in columns 7 and 8, respectively. Generated $T_1$ images from column 3 with pix2pix and one2one models are in columns 9 and 10, respectively. In columns 3-6, the score under each image is its PSNR and SSIM score compared with the original image. In column 7-10, the scores under each image are the PSNR and SSIM score differences between input $x$ and $x+dx$ for both models. For example, to compare model sensitivity on $T_1 \rightarrow T_2$ direction, $x$ is the column 1, $x+dx$ is the column 5. The model sensitivity for the pix2pix model is the score difference between columns 3 and 7. The model sensitivity for the one2one model the score difference between columns 4 and 8.}
\label{fig:long}
\label{fig:onecol}
\end{figure*}

\subsection{Map $\leftrightarrow$ aerial photo}

Table 2 and Figure 6 show the model performance comparison between one2one model and pix2pix model on bidirectional aerial and map image translation. In the direction of aerial photo$\rightarrow$ image translation is many-to-one. As shown in Table 2 and Figure 6 upper part, pix2pix produces better result than one2one by 3\%, 10.5\%, 9,6\% in PSNR,SSIM and L1 individually. In the direction of map$\rightarrow$ aerial photo, as shown in Table 2 and bottom part of Figure 6, the one2one model outperform the pix2pix model by 3\% in SSIM and 2\% in PSNR.

\subsection{MRI image synthesis on BRATS}

We conduct the experiments based on the BraTS 2018 dataset \cite{brats1}, which contain ample multi-institutional routine clinically-acquired pre-operative multimodal MRI scans of glioblastoma (GBM/HGG) and  lower grade glioma (LGG) images. There are 285 3D volumes for training and 66 3D volume for test. The $T_1$ and $T_2$ images are selected for our bi-directional image synthesis. All the 3D volumes are preprocessed to one channel image of size 256 x 256 x 1. In all tables, all results are averaged across all splits as in~\cite{brats1}. As shown in Table 5(a), on the $T_1 \rightarrow T_2$ image synthesis direction, our one2one model outperforms the pix2pix model on PSNR by 13.6\%. The qualitative result is shown in columns 3 and 4 in Figure 9. On the $T_2 \rightarrow T_1$ image synthesis direction, our one2one model outperforms the pix2pix model on PSNR by 11.6\%.  The qualitative result is shown in columns 5 and 6 in Figure 9.

\begin{table*}[t]
\begin{center}
\begin{tabular}{|l|l|l|l|l|l|l|l|l|}
\hline
Direction & Method & (a) & L1$\downarrow$ & PSNR$\uparrow$ & SSIM$\uparrow$ & (b) & d$\|$PSNR$\|$ $\uparrow$ & d$\|$SSIM$\|$ $\uparrow$\\
\hline\hline
$T_1 \rightarrow T_2$ &  {pix2pix} &&0.042&26.53&0.871 & &2.17&0.018 \\
$T_1 \rightarrow T_2$ &  {one2one} &&\textbf{0.039}&\textbf{29.23}&\textbf{0.875}& &\textbf{3.01}&\textbf{0.020}\\

\hline
$T_2 \rightarrow T_1$ &  {pix2pix} &&0.051&27.78&0.872&&4.51&0.034\\
$T_2 \rightarrow T_1$ &  {one2one} &&\textbf{0.048}&\textbf{30.99}&\textbf{0.876}&&\textbf{4.93}&\textbf{0.036}\\
\hline\end{tabular}
\end{center}
\caption{(a) Image synthesis performance and (b) model sensitive analysis on MRI T1 and T2 images from BraTs dataset\cite{brats1}. L1 is the smaller the better. The difference of PSNR and SSIM are the larger the more sensitive. All the metrics are averaged on 10230 1-channel 2D images. }
\end{table*}

\section{Model sensitivity analysis}
To measure the model sensitivity, we add a perturbation $dx$ to the input image $x$, then measure the change of the output, $dy$. In our experiment on BraTs dataset shown in Figure.9, on the $T_1 \rightarrow T_2$ direction, the input image with perturbation $x+dx$ is the generated $T_1$ images from $T_2$ with the pix2pix model (see colunum 5 in Figure.9), on the $T_2 \rightarrow T_1$ direction, the input image with perturbation $x+dx$ is the generated $T_2$ images from $T_1$ with the pix2pix model (see column 3 in Figure.9).

In order to compare the performance of pix2pix and one2one on both tasks $A$ and $B$, we need to train 3 models in total: pix2pix for task $A$ (pix2pixA), pix2pix for task $B$ (pix2pixB) and a one2one model for both tasks $A$ and $B$ (one2one). To compare the model sensitivity between pix2pixA and one2one for task $A$, we follow four steps. 
\begin{enumerate}
    \item For an image pair $(x_{i},y_{i})/(T_1,T_2)$, we pass $y_i/T_2$ to pix2pixB as input to generate $x_i+dx_i/ T_1'(pix2pix)$, which adds a perturbation to $x_i/T_1$. 
    \item We input $x_i/T_1$ to the pix2pixA and one2one models, obtaining the corresponding outputs $y_i' /T_2'(pix2pix)$ and $y_i'/T_2'(one2one)$, respectively.
    \item We input $x_i+dx_i$ to the pix2pixA and one2one models obtaining the corresponding outputs ${(y_i+dy_i)}' /T_2''(pix2pix)$ and ${(y_i+dy_i)}'/T_2''(one2one)$, respectively.
    \item For both models, we use a predefined evaluation metric $E$ (for example PSNR and SSIM) to evaluate $y_i^{\prime}$and ${(y_i+dy_i)}^{\prime}$ and get the scores $E{y_i^{\prime}}$ and $E{(y_i+dy_i)^{\prime}}$, respectively. So, the change of the output is measured by $d\|E\| = | E{(y_i+dy_i)^{\prime}}-E{y_i^{\prime}}|$.
\end{enumerate}

The model with a larger change of the output due to perturbation $dx_i$ is more sensitive, and vice versa. Similarly, we can compare the model sensitivity between pix2pixB and one2one for task $B$ by swapping the $x_i$ and $y_i$ in the above steps. %Another difference is in the first step, we pass $x_i$ to pix2pixA as input to generate $y_i+dy_i$.

As shown in Table 5(b) on the $T_1 \rightarrow T_2$ image synthesis direction, our one2one model is more sensitive than pix2pix model, improving PSNR by 38.7\%! The qualitative result is shown in column 7 and 8 in Figure 9. On the $T_2 \rightarrow T_1$ image synthesis direction, our one2one model is more sensitive than pix2pix model, improving PSNR by 9.3\%. The qualitative results are shown in columns 9 and 10 in Figure 9.

For the cityscapes dataset, we use the mean class IOU to measure the change of output for photo $\rightarrow$ labels direction and ``FCN score" to measure the change of output for labels $\rightarrow$ photo direction. In Table 3 and figure 7, D(CLASS IOU) is the absolute value difference of IOU score for the photo$\rightarrow$labels direction and FCN score for label$\rightarrow$photo direction between one2one and pix2pix.

For the Google Maps data set, we use the structural similarity index (SSIM), peak signal to noise ratio (PSNR) and L1 distance to measure the change of output from both directions. In Table 4 and Figure 8, the dL1, dPSNR and dSSIM is the absolute value of the difference between one2one and pix2pix.

For the cityscapes dataset, according to Table 3, one2one model is more sensitive than pix2pix by 6\% in the label $\rightarrow$ photo direction and 5\% in the photo $\rightarrow$ label direction and Figure 8 illustrates qualitative sensitivity analysis. 

For the maps dataset, according Table 4, one2one model is more sensitive than pix2pix by 2\% in PSNR and 14\% in L1 for the aerial $\rightarrow$ map direction. The one2one model is more sensitive than pix2pix by 3\% in L1, 2\% in PSNR and 4.3\% in SSIM in the map $\rightarrow$ aerial direction. Figure 8 illustrates qualitative sensitivity analysis.

In summary, the one2one model is more sensitive than pix2pix models on all the three datasets.

\section{Conclusion}
We have presented an approach for learning one U-Net for both forward and inverse image-to-image translation. The experiment results and model sensitivity analysis results are consistent to verify the one-to-one mapping property of the self-inverse network. In future, we will further explore the theoretical aspect of the self-inverse network learning.

{\small
\bibliographystyle{ieee}
\bibliography{egpaper_final}
}

\end{document}